\documentclass{article}
\PassOptionsToPackage{numbers}{natbib}
\usepackage[final]{neurips_2026}
\makeatletter
\renewcommand{\@noticestring}{Accepted at the 9th Workshop on Machine Learning and the Physical Sciences (ML4PS 2026).}
\makeatother
\usepackage[utf8]{inputenc} 
\usepackage[T1]{fontenc}    
\usepackage{hyperref}       
\usepackage{url}            
\usepackage{booktabs}       
\usepackage{amsfonts}       
\usepackage{nicefrac}       
\usepackage{microtype}      
\usepackage{xcolor}         
\usepackage{amsmath}        
\usepackage{mathrsfs}       
\usepackage{graphicx}
\usepackage{float}
\usepackage{cancel}

\title{PINN-Slicer: Exploiting Rotational Symmetry to Solve the Photon Orbit Equation Around a Black Hole}
\author{%
  Claire David\thanks{Part of this work was carried out while the author was at AIMS South Africa.} \\
  Independent \\
  \And
  Tlotlo Oepeng \\
  De Beers Group \\
  \And
  Harrison B. Prosper \\
  Florida State University \\
  Tallahassee, USA \\
}
\begin{document}
\maketitle
\begin{abstract}
Physics-Informed Neural Networks (PINNs) typically struggle with long-horizon domains, often requiring the network to be retrained as the domain grows. We present PINN-Slicer, a method for second-order autonomous ordinary differential equations, $\ddot{u} = G(u, \dot{u})$, which exploits the invariance of this class of equations under translations of the independent variable. A \emph{single} PINN, embedded within a solution \emph{ansatz} that enforces the initial conditions explicitly, is trained on a thin slice of the independent variable using Sobol-sampled initial conditions. The PINN is then deployed recursively, slice-by-slice, across the full domain. We demonstrate PINN-Slicer on the nonlinear equation that describes photon orbits around a black hole, and reach a point-by-point root-mean-square error (RMSE) below $10^{-3}$ in $u = r_s/r$ (the inverse radial coordinate in units of the Schwarzschild radius) relative to a finite-difference reference solution. A key advantage of PINN-Slicer is that a single trained network generalizes to any orbital initial condition within the training range, without retraining. In addition, PINN-Slicer can cope with solution domains that may not be known \emph{a priori}.
\end{abstract}

\section{Introduction}

The work reported here was motivated by the following questions: (i) can a Physics-Informed Neural Network (PINN) \cite{Raissi2019} accurately model the solutions of an ordinary differential equation (ODE) when the physical domain of the solutions may be unknown, and (ii) can it provide an accurate surrogate for solutions that are differentiable with respect to the initial conditions \cite{SUN2020112732}? Photon orbits around a black hole can escape to infinity or terminate at the Schwarzschild radius, and can be sensitive to the initial conditions \cite{hartle2021gravity}. They therefore provide a well-studied benchmark for these questions. PINNs, however, remain challenging to train \cite{wang2020pinnsfail}: models that perform well on simple problems can fail on slightly more complex ones \cite{krishnapriyan2021failure}. Proposed remedies include domain decomposition \cite{djagtap2020, jagcpinns}, curriculum and sequential training \cite{krishnapriyan2021failure, wight2020}, and modifications to the optimization objective \cite{wang2020pinnsfail}.

In this paper, we present PINN-Slicer,\footnote{Code is available at \url{https://github.com/soot-bit/pinnslicer/}.} a method for second-order autonomous ODEs. PINN-Slicer is trained on a single segment\footnote{We use the terms \emph{slice} and \emph{segment} interchangeably.} of the independent variable and applied recursively to construct the full solution, a strategy that resembles that of Wang and Perdikaris \cite{wang2023}. However, our method incorporates initial conditions explicitly thanks to an \emph{ansatz} from the theory of connections \cite{Mortari2017}, so that our loss is based solely on the ODE residuals.

\section{The photon orbit equation}

\begin{figure}
    \centering
    \includegraphics[width=0.75\linewidth]{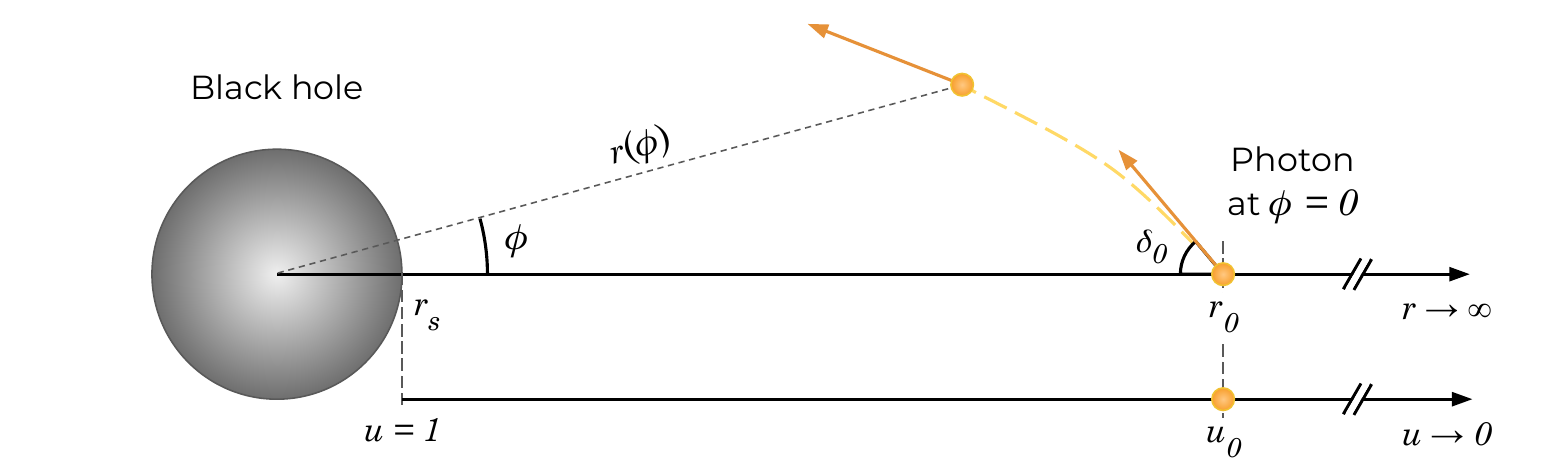}
    \caption{Coordinate system: a photon emitted at $(r_0, \delta_0)$ from a
black hole of Schwarzschild radius $r_s$ follows a trajectory
parametrized by radius $r$ (or $u = r_s/r$) and azimuthal angle
$\phi$, measured from the initial radial direction (set to the
horizontal here without loss of generality). 
    }
    \label{fig:coord_init_cond}
\end{figure}

We focus on the equation~\cite{hartle2021gravity, atwater1974introduction}
\begin{align}
\ddot{u} & \,= \frac{3}{2} \, u^2 - u, 
\label{eq:ode_u}
\end{align}
which describes the trajectory of a photon around a black hole in the
Schwarzschild metric. Here $u = r_s/r$ is the inverse radial
coordinate in units of the Schwarzschild radius $r_s$, and the
overdot ($\ddot{u}$) denotes differentiation with respect to the azimuthal angle
$\phi$. Figure~\ref{fig:coord_init_cond} illustrates the geometry,
together with the initial conditions $(r_0, \delta_0)$ at $\phi = 0$.
These initial conditions can also be expressed in the bounded variables
\begin{align}
  u_0 = \frac{r_s}{r_0}, \qquad \qquad  v_0 = \dot{u}(0),
\end{align}
where $v_0$ follows from $\delta_0$ via the initial radial velocity (see Appendix~\ref{app:v0}). 
In this context, the physical range of $\phi$ is not known \emph{a priori}: depending on the initial conditions, the photon either escapes to infinity, is
absorbed by the black hole, or orbits indefinitely (the special case of an unstable circular orbit on the photon sphere\,\cite{hartle2021gravity}). The extent of $\phi$ for which a solution exists is, therefore, determined only by solving the equation itself.

\section{Methodology}

\subsection{PINN}
PINNs are typically trained by minimizing a composite loss combining a data
term $\mathcal{L}_{\text{data}}$, a physics term
$\mathcal{L}_{\text{ode}}$ embedding the governing physics via the
residual of the ODE, and a boundary/initial-condition term
$\mathcal{L}_{\text{bc}}$ \citep{Raissi2019},
\begin{align}
  \mathcal{L}_{\text{total}}(\omega) = w_{\text{data}}\mathcal{L}_{\text{data}}(\omega)
  + w_{\text{ode}}\mathcal{L}_{\text{ode}}(\omega) + w_{\text{bc}}\mathcal{L}_{\text{bc}}(\omega),
\end{align}
where $\omega$ are the network's trainable parameters. The
coefficients $w_{\text{data}}, w_{\text{ode}}, w_{\text{bc}}$ are
additional hyperparameters weighting the relative contribution of each
term.
In our setting there are no observed data to fit, so
$\mathcal{L}_{\text{data}}$ is absent. Instead, the PINN is trained on $N = 2^{16}$ points $(\phi, u_0, v_0)_i$, $i = 0,\ldots,N-1$, sampled using Sobol
sampling on the domain $[0, \Delta\phi] \times [0.10, 0.99] \times [-1, 1]$.
We chose Sobol sampling over uniform sampling because Sobol sequences
are quasi-random and cover the sampling space more evenly than pure
random sampling. 
As the equation has no spatial boundary, only an initial condition at
$\phi=0$, the boundary term reduces to an initial-condition term
$\mathcal{L}_{\text{ic}}$. The composite loss becomes
\begin{align}
  \mathcal{L}_{\text{total}}(\omega) = w_{\text{ode}}\,\mathcal{L}_{\text{ode}}(\omega)
  + w_{\text{ic}}\,\mathcal{L}_{\text{ic}}(\omega).
\end{align}
This still leaves two coefficients, $w_{\text{ode}}$ and
$w_{\text{ic}}$, which need to be carefully chosen. Different terms in the loss function often converge at different rates, complicating training \citep{wang2020pinnsfail}, and 
PINN training can fail unpredictably as problem complexity increases only slightly~\citep{krishnapriyan2021failure}.

\subsection{Solution \textit{ansatz}}
Instead of a soft constraint on the initial conditions as an
additional term in the loss function, we adopt an \textit{ansatz} from the theory of connections (ToC)
\citep{Mortari2017},
\begin{align}
  u(\phi, u_0, v_0) &= u_0 + g_\omega(\phi, u_0, v_0) - g_\omega(0, u_0, v_0)
    + \phi\left[v_0 - \dot g_\omega(0, u_0, v_0)\right], \nonumber\\
  \dot u(\phi, u_0, v_0) &= v_0 + \dot g_\omega(\phi, u_0, v_0) - \dot g_\omega(0, u_0, v_0),
  \label{eq:constrained}
\end{align}
which incorporates the initial conditions directly and includes them
as inputs to $g_\omega$, a feedforward network with trainable
parameters $\omega$; $\dot g_\omega$ denotes its derivative with
respect to $\phi$. The network is a fully connected architecture with
3 hidden layers of 75 neurons each and sine activation functions. We
did not perform systematic hyperparameter optimization.

Thanks to the \textit{ansatz} in Equation~\ref{eq:constrained}, the
initial conditions are satisfied explicitly, by construction, rather
than through a loss penalty. This removes $\mathcal{L}_{\text{ic}}$
and its coefficient, leaving a loss with one term and no coefficient
to tune.

\subsection{Domain slicing strategy}

\begin{figure}
    \centering
    \includegraphics[width=\linewidth]{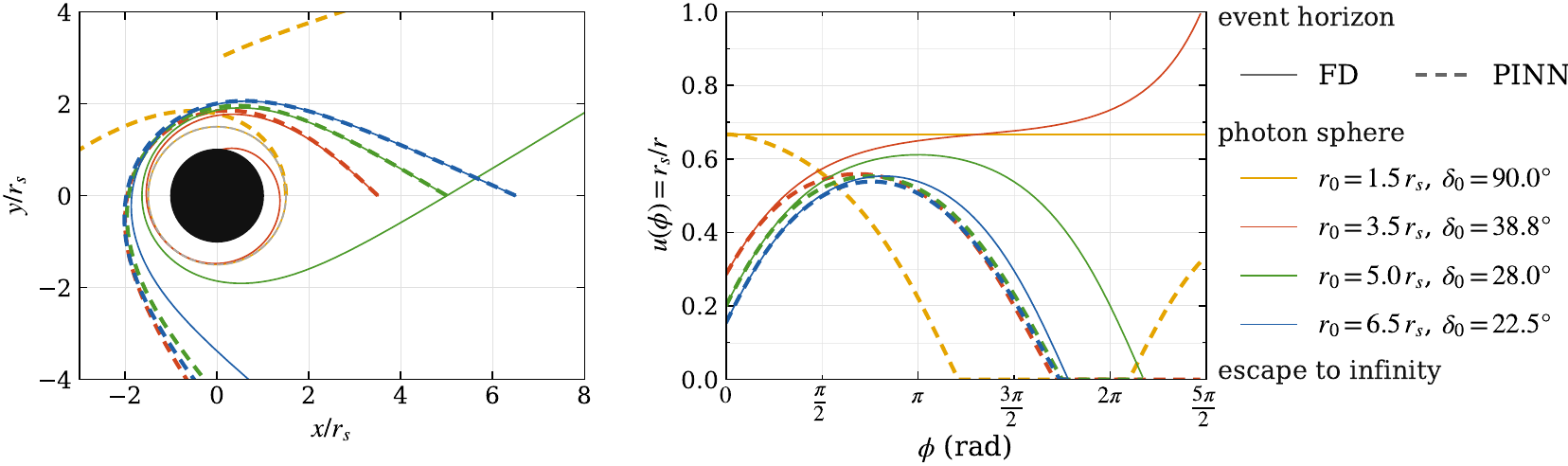}
    \caption{Four orbits computed with the finite-difference solver (FD)
and the PINN, trained and deployed over $[0, 5\pi/2]$,
shown in the $(x,y)$ plane (left) and the $(\phi, u)$ plane (right).
The PINN fails over its own training domain.}
    \label{fig:failure_demo_orbits}
\end{figure}
We use a custom fourth-order finite-difference (FD) solver to compare with
the PINN solutions (Appendix~\ref{app:fd}). A first deployment of our PINN, trained over the range $[0,5\pi/2]$, leads to a catastrophic failure, as seen in
Figure~\ref{fig:failure_demo_orbits}. 
Appendix~\ref{app:boomerang} shows a further example using boomerang orbits, which are highly sensitive  to small perturbations and therefore a particularly stringent test of the method. 
Such failures are not a surprise, as PINNs
are known to fail even for relatively simple problems~\citep{krishnapriyan2021failure}.

Various mitigating strategies show
that PINNs can benefit from decomposing the computational domain
(cPINN~\citep{jagcpinns}, XPINN~\citep{djagtap2020}). Inspired by
this, we propose a recursive method exploiting the key observation
that our ODE is \emph{autonomous}. In other words, the coordinate
system can be rotated to a different starting azimuthal angle. 
\begin{figure}
    \centering
    \includegraphics[width=\linewidth]{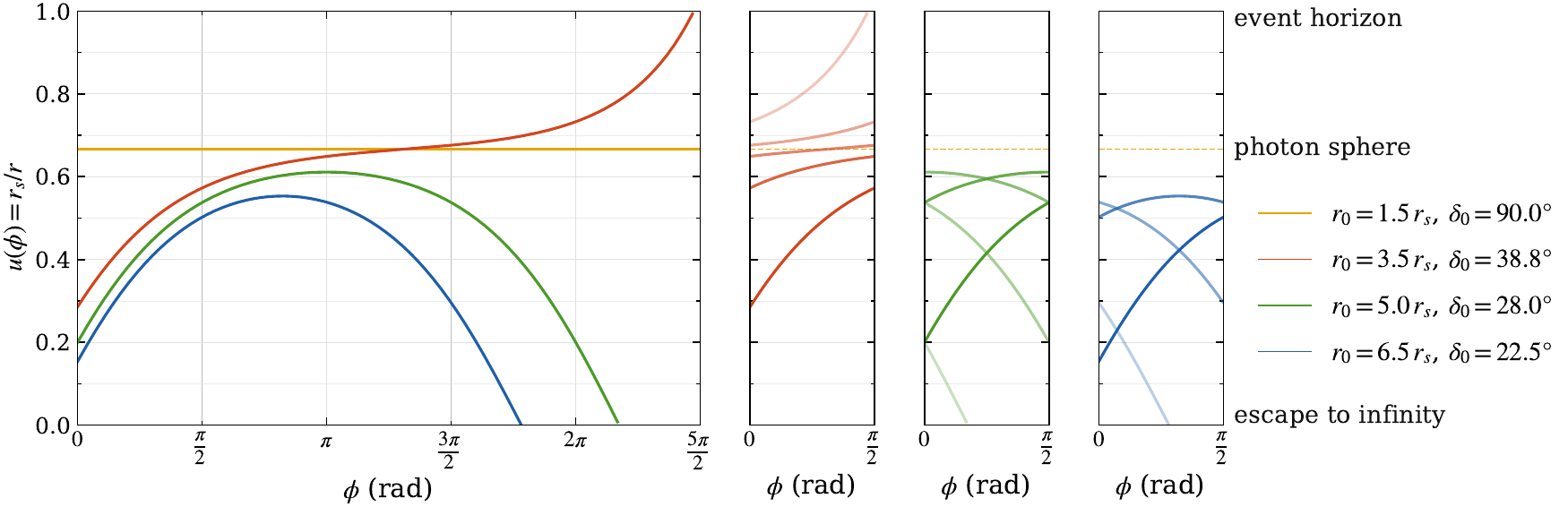}
    \caption{Illustration of the slicing strategy. Four orbits
are computed with the FD solver over the full $\phi$ domain (left). For
three of these orbits, the segments of width $\Delta\phi$ are
overlaid on the common slice $[0, \Delta\phi]$ (right).}
    \label{fig:fd-segment-overlay}
\end{figure}
Therefore, the PINN is trained and deployed only over a thinner slice
$[0, \Delta\phi]$. Our recursive algorithm stitches together
successive $\Delta\phi$ segments: the solution $u(\Delta\phi)$ and
its derivative $v(\Delta\phi)$ at the end of one segment become the
initial conditions $(u_0, v_0)$ for the next. 
The procedure is repeated until
the desired $\phi$ coverage is reached. 
Figure~\ref{fig:fd-segment-overlay} shows
three orbits in the $(\phi, u)$ plane, together with three side
panels displaying, for each orbit, its segments overlaid on the
common slice $[0, \Delta\phi]$. For illustration purposes, a large segment of
$\Delta\phi = \pi/2$ is used; the method's actual training segment
is smaller (see Section~\ref{sec:results}). 

\section{Results}
\label{sec:results}

PINN-Slicer was trained on an NVIDIA GeForce RTX 5090 for
$1$ million iterations at $452$ iterations/second, completing in
37 minutes. We trained several PINNs over different $\Delta\phi$ segments: 0.05, 0.1, 0.2, 0.4, 0.8, and 1.6 radians. 
The lowest mean validation loss, $2.07\times10^{-10}$, is achieved at $\Delta\phi = 0.1$ (see Appendix~\ref{app:loss-vs-dphi} for the full comparison across segments).

For $\Delta\phi = 0.1$, the largest point-by-point root-mean-square error (RMSE) among the four orbits is $4.94\times10^{-4}$ in $u$ and $5.96\times10^{-4}$ in $v$, computed over the $\phi$ points where both solutions exist.
Over a $41 \times 41$ grid of initial conditions covering the training region (1,642 comparable orbits), every orbit has RMSE$_u < 10^{-3}$ for $\Delta\phi = 0.1$ (Appendix~\ref{app:rmse-vs-ic}).

\begin{figure}[h]
    \centering
    \includegraphics[width=1\linewidth]{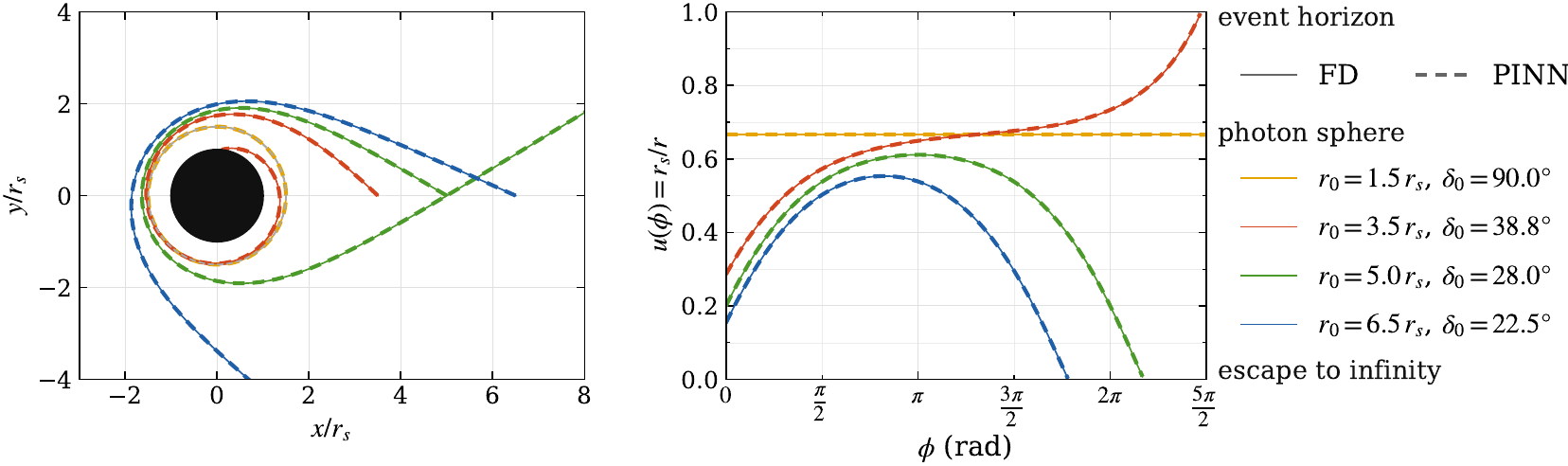}
    \caption{Solutions of PINN-Slicer, trained over the slice $[0, \Delta\phi]$, with $\Delta\phi = 0.1$, and recursively deployed over 
    the full domain of $[0, 5\pi/2]$, closely
matching the FD solver reference solutions.}
    \label{fig:fdpinn-slicer010}
\end{figure}
\begin{table}[h]
\centering
\caption{Per-orbit results for the PINN trained with a
$\phi$-slice of $[0, 0.1]$: RMSE relative to the FD solution, the orbit type, and the physical range of the solution domain.}
\label{tab:rmse-solution-domain}
\begin{tabular}{cccccccc}
\toprule
$r_0$ & $\delta$ & $u_0$ & $v_0$ & RMSE$_u$ & RMSE$_v$ & $\phi_{\text{reached}}$ & Type \\
\midrule
1.50 & 90.00 & 0.6667 & 0.00 & $2.44\times10^{-4}$ & $2.45\times10^{-4}$ & 7.854 & photon sphere \\
3.50 & 38.82 & 0.2857 & 0.30 & $4.94\times10^{-4}$ & $5.96\times10^{-4}$ & 7.756 & inward spiral \\
5.00 & 27.99 & 0.2000 & 0.34 & $1.06\times10^{-4}$ & $5.61\times10^{-5}$ & 6.811 & escaping \\
6.50 & 22.50 & 0.1538 & 0.34 & $2.07\times10^{-5}$ & $1.76\times10^{-5}$ & 5.590 & escaping \\

\bottomrule
\end{tabular}
\end{table}
\section{Discussion}
Our study demonstrates a successful deployment of a PINN over a long-horizon domain, achieved through recursion over the independent variable and reaching an RMSE below $10^{-3}$. PINN-Slicer recovers both the solution and its physical domain. We believe this method generalizes to any second-order \emph{autonomous} ODE. Work is underway to apply and test PINN-Slicer on other such equations across various branches of physics.

\bibliographystyle{unsrtnat}
\bibliography{pinn_ml4ps2026}
\appendix

\section{Derivation of the initial condition \texorpdfstring{$\boldsymbol{v_0(\delta_0)}$}{v0(delta0)}}
\label{app:v0}

We recall the two initial conditions of our 2nd order ODE are expressed in terms of the bounded variables:
\begin{align}
  u = \frac{r_s}{r} \qquad \qquad  v = \dot{u},
\end{align}
where $u$ is the inverse radial coordinate $u = r_s/r$, with $r_s$ the Schwarzschild radius. Without loss of generality, we take $\phi = 0$ to be the initial value of the azimuthal coordinate:
\begin{align}
  u_0 = \frac{r_s}{r(\phi=0)}, \qquad \qquad v_0 = \dot{u}(\phi=0).
\end{align}

\begin{figure}[h]
    \centering
    \includegraphics[width=0.9\linewidth]{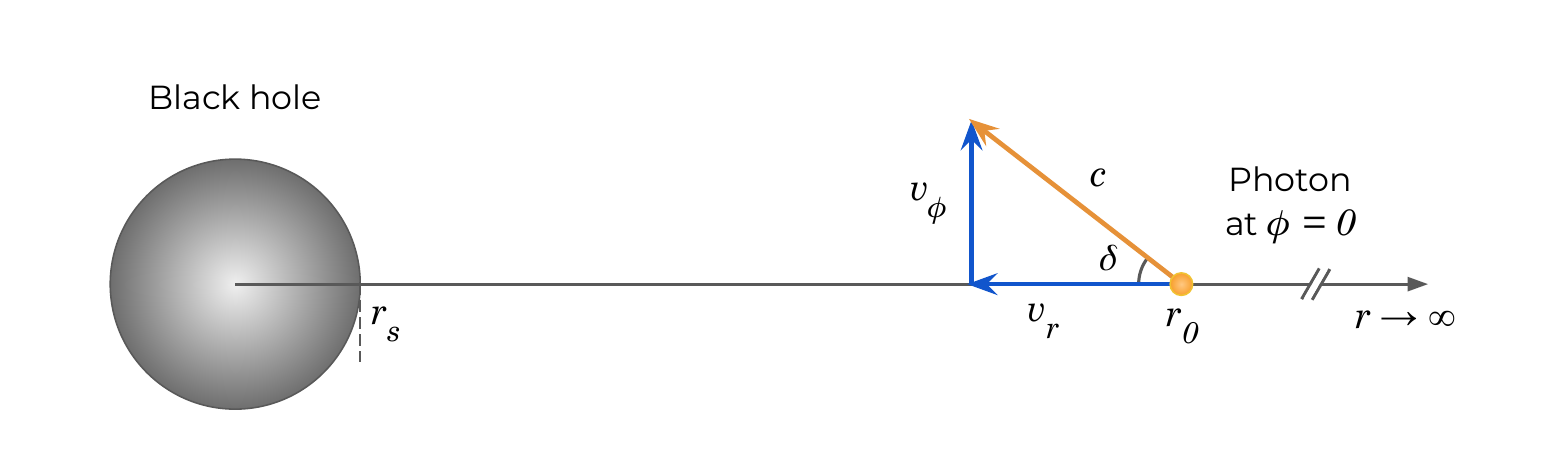}
    \caption{Definition of photon emission angle $\delta$ and its relationship to the radial and angular photon velocities.}
    \label{fig:photon-v-coordinates}
\end{figure}

However, one can relate the derivative to the more intuitive emission angle $\delta_0$, as shown in Figure~\ref{fig:photon-v-coordinates}, between the initial photon direction and the radial vector pointing
towards the center of mass from the initial point, $(r=r_0, \phi=0)$.

To compute $v_0$ in terms of this emission angle $\delta_0$, we start with the metric for null geodesics (the spacetime paths followed by massless particles),
\begin{align}
    ds^2 & = f (c dt)^2 - (f^{-1} dr^2 + r^2 \, d\phi^2) = 0,
\end{align}
which we rewrite as follows,
\begin{align}
    c^2 & = f^{-1} \left(\frac{dr}{d\tau}\right)^2 + r^2 \, \left(\frac{d\phi}{d\tau}\right)^2,\quad d\tau = f^{1/2}\,dt,
\end{align}
where $f = 1 - u$ and $d\tau$ is the proper time, that is, the time recorded by an observer at spatial coordinate $(r, \phi) \equiv (u, \phi)$. The expression above is an instance of Pythagoras's theorem: the first
term on the right is the square of the radial speed, $v_r$, of the photon while the second term is the square of the angular speed,~$v_\phi$. The sum $v_r^2 + v_\phi^2$ necessarily equals the square of
the speed of light in vacuum. The angle of interest is given by
\begin{align}
    \tan \delta & = \frac{v_\phi}{v_r},\\
        & = \pm \frac{r d\phi/d\tau}{f^{-1/2}dr/d\tau},\nonumber\\
        & = \pm r \sqrt{1-u} \, / \,  \frac{dr}{d\phi}, \nonumber\\
        &  = \pm u \sqrt{1-u} \, / \,  \frac{du}{d\phi}, \\
        \rightarrow v(\phi) \equiv \frac{du}{d\phi} & = \pm \frac{u \sqrt{1-u}}{\tan \delta}, \quad\text{valid for } u < 1.
\end{align}
At $\phi = 0$, and taking the positive sign to be consistent with Figure~\ref{fig:photon-v-coordinates} we find
\begin{align}
    v_0 & = \frac{u_0 \sqrt{1-u_0}}{\tan \delta} \,.
    \label{eq:v0fdelta0}
\end{align}

\section{Custom finite-difference solver}
\label{app:fd}

We developed a custom finite-difference solver, giving us full
control over the accuracy of the reference solutions, specifically for the equation
\begin{align}
    \ddot{u} = G(u), \quad G(u) = \frac{3u^2}{2} - u,
\end{align}
exploiting its algebraic structure to compute exact derivatives, here
up to fourth order. Its accuracy can be systematically extended to
higher order.

The 2nd-order ODE is coded within the solver as two coupled 1st-order
equations
\begin{align}
    D u & = v,\\
    D v & = G,
\end{align}
where $D$ is the derivative operator, $d/d\phi$.

The solution at $u(\phi+h)$ can be computed using the Taylor series
expansion:
\begin{align}
    u(\phi+h) & = u + Du \, h + \frac{D^2 u \, h^2}{2!} + \frac{D^3 u \, h^3}{3!} + \frac{D^4 u \, h^4}{4!} + \mathscr{O}(h^5), \\
        & = u + v \, h + \frac{G \, h^2}{2} + \frac{DG \, h^3}{6} + \frac{D^2 G \, h^4}{24} + \mathscr{O}(h^5).
\end{align}

Applying the operator $D$ to $u(\phi + h)$ yields
\begin{align}
    v(\phi+h) & = v + G \, h + \frac{DG \, h^2}{2} + \frac{D^2 G \, h^3}{6} + \frac{D^3 G \, h^4}{24} + \mathscr{O}(h^5).
\end{align}

Because of the simplicity of the function $G(u) = \frac{3u^2}{2} - u$,
the quantities $DG$, $D^2G$, and $D^3G$ are readily computed:
\begin{align}
    DG & = D \left[ \frac{3 u^2}{2} - u \right], \nonumber\\
            & = 3 u Du - Du,\nonumber\\
            & = Q v, \quad\text{ where } Q \equiv 3 u - 1,\\
            \nonumber\\
    D^2G & = (D Q) v + Q (D v), \\
         & = 3 v^2 + Q G,\\ \nonumber\\
    D^3G & = (9 G + Q^2) v.
\end{align}

We thus arrive at explicit formulae for $u(\phi+h)$ and $v(\phi+h)$
that depend solely on the known quantities $u(\phi)$, $v(\phi)$, and
$G(\phi)$. The solution is arrived at through iterative application
of these formulae.

\section{Boomerang orbits}
\label{app:boomerang}

\begin{figure}[H]
    \centering
    \includegraphics[width=\linewidth]{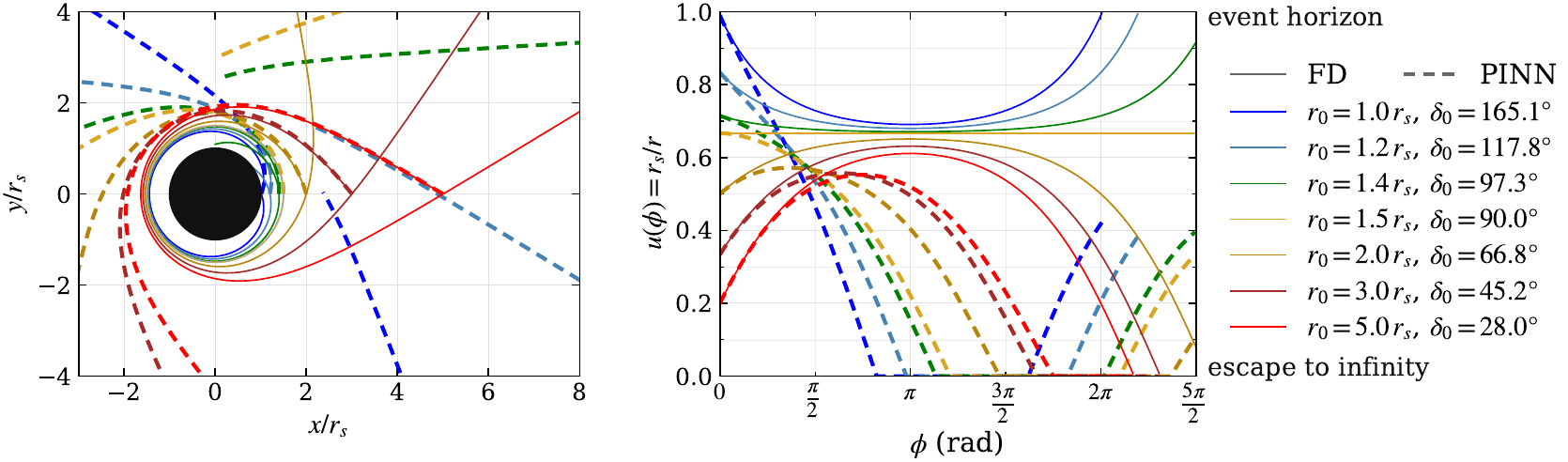}
    \caption{Seven boomerang orbits computed with the finite-difference
solver (FD) and the PINN, trained and deployed over
$\Delta\phi$ over $[0, 5\pi/2]$, shown in the $(x,y)$ plane (left) and
the $(\phi, u)$ plane (right). The PINN fails over its own training
domain.}
    \label{fig:pinn-failure-boomerang}
\end{figure}

\begin{figure}[H]
    \centering
    \includegraphics[width=1\linewidth]{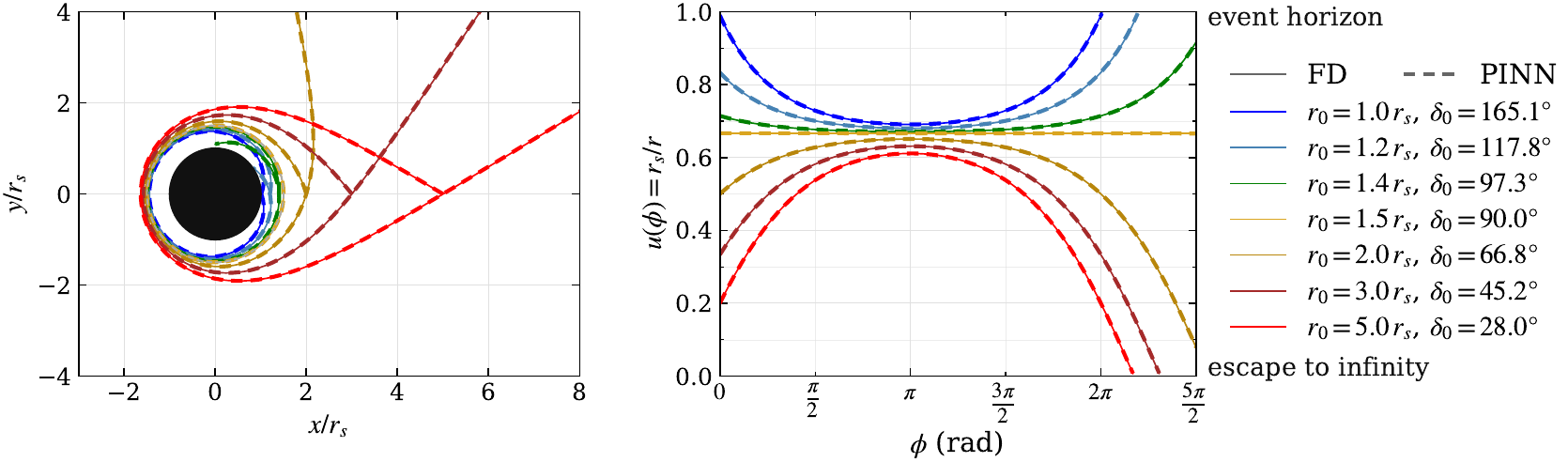}
    \caption{Seven boomerang orbits computed with the finite-difference
solver (FD) and PINN-Slicer, shown in the $(x,y)$ plane (left) and the $(\phi, u)$ plane (right). PINN-Slicer was trained over $\Delta\phi = 0.1$ and deployed over the domain $[0, 5\pi/2]$. PINN-Slicer accurately reproduces the boomerang orbits.}
    \label{fig:pinnslicer-boomerang}
\end{figure}

A boomerang orbit is a special case in which the photon's trajectory
encircles the black hole before passing back through its initial
point of emission, for a precise and specific pair $(u_0, v_0)$. 
This makes boomerang orbits a stringent test of solver accuracy, as any small deviation is immediately visible: the trajectory simply
misses the starting point. 
Figure~\ref{fig:pinn-failure-boomerang}
shows the failure of a PINN trained and deployed over a large domain.
With PINN-Slicer, the boomerang orbits are accurately reproduced (Figure~\ref{fig:pinnslicer-boomerang}). 

\section{Validation loss across training segments}
\label{app:loss-vs-dphi}

\begin{figure}[H]
    \centering
    \includegraphics[width=0.5\linewidth]{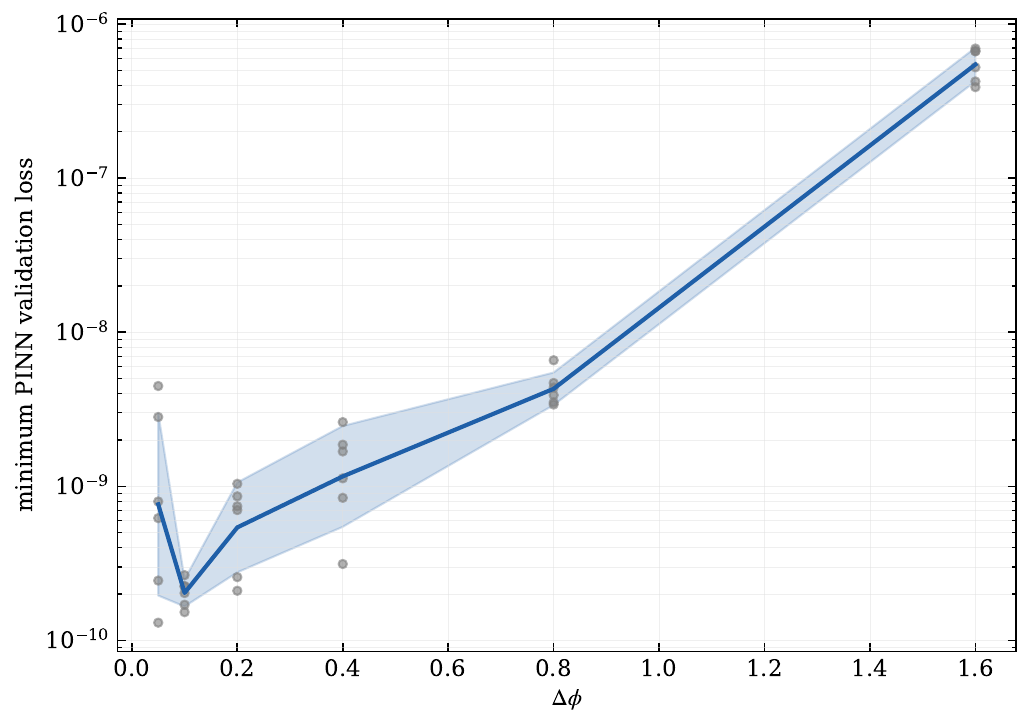}
    \caption{Average validation loss (at the end of training) as a function of $\Delta\phi$ for 6 training runs, each associated with a different random number seed. The solid line shows the mean over these runs. The shaded band is the standard deviation. The grey points
    are the individual final average losses. The lowest average loss is achieved at
    $\Delta\phi = 0.1$.}
    \label{fig:loss-vs-dphi}
\end{figure}

\section{Additional FD--PINN comparisons}
\label{app:xy-uphi}

\begin{figure}[h]
    \centering
    \includegraphics[width=\linewidth]{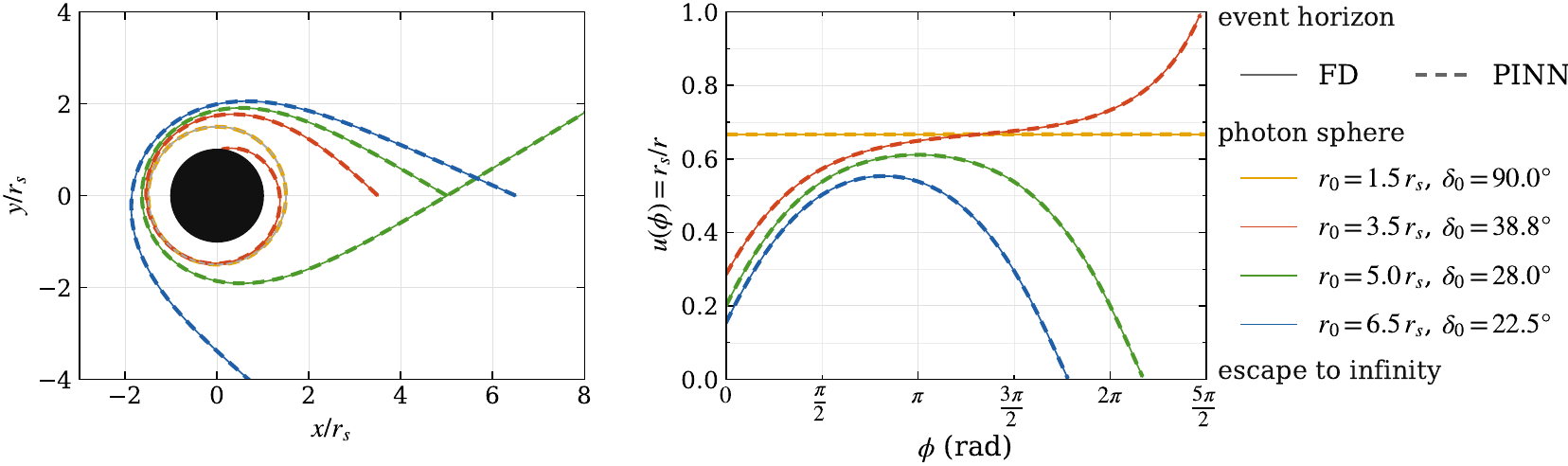}
    \caption{PINN-Slicer solutions, trained with $\Delta\phi = 0.05$, are in
excellent agreement with the FD reference solutions over the full
domain, despite the slightly larger final average validation loss
(Figure~\ref{fig:loss-vs-dphi}).}
    \label{fig:fdpinn005}
\end{figure}

\begin{figure}[h]
    \centering
    \includegraphics[width=\linewidth]{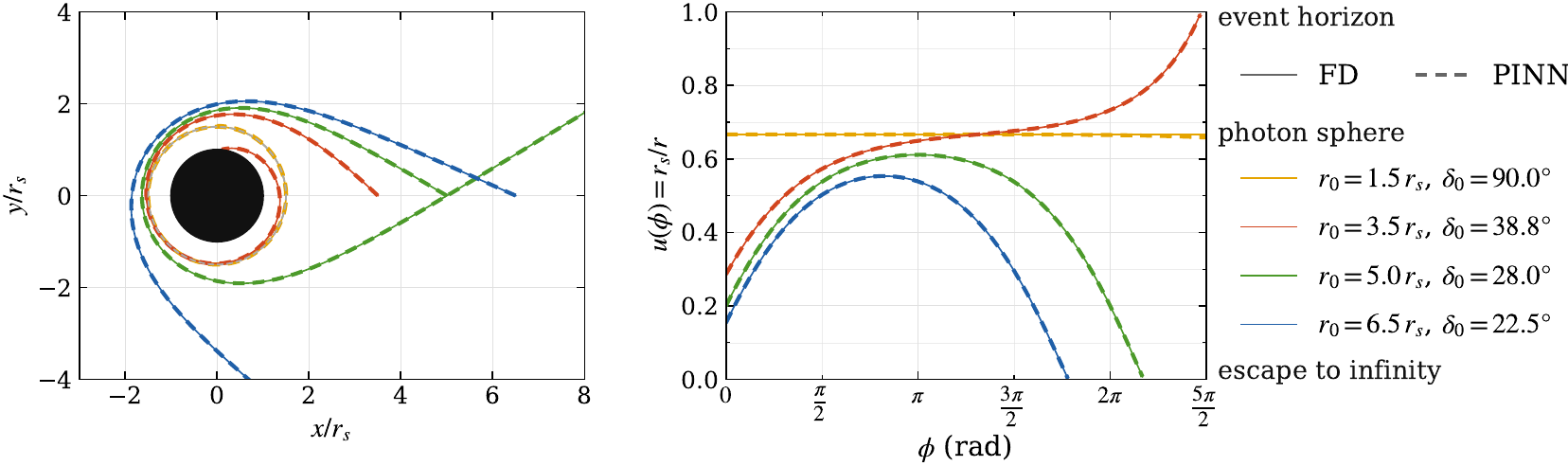}
    \caption{PINN-Slicer solutions, trained with $\Delta\phi = 0.2$, exhibit a very slight deviation near the end of the domain, mostly visible on the inward-spiral orbit (red).}
    \label{fig:fdpinn020}
\end{figure}

\begin{figure}[h]
    \centering
    \includegraphics[width=\linewidth]{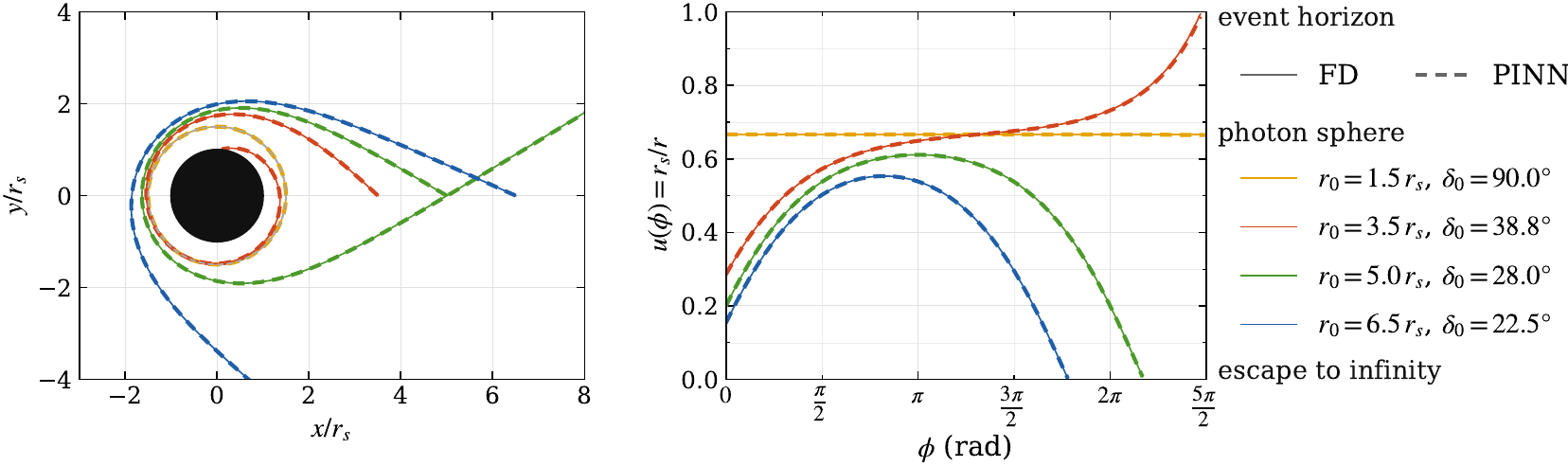}
    \caption{FD and PINN-Slicer solutions for $\Delta\phi = 0.4$. The
    PINN begins to deviate from the FD reference near the end of the
    domain, most visibly for the inward-spiral orbit.}
    \label{fig:fdpinn040}
\end{figure}

\begin{figure}[h]
    \centering
    \includegraphics[width=\linewidth]{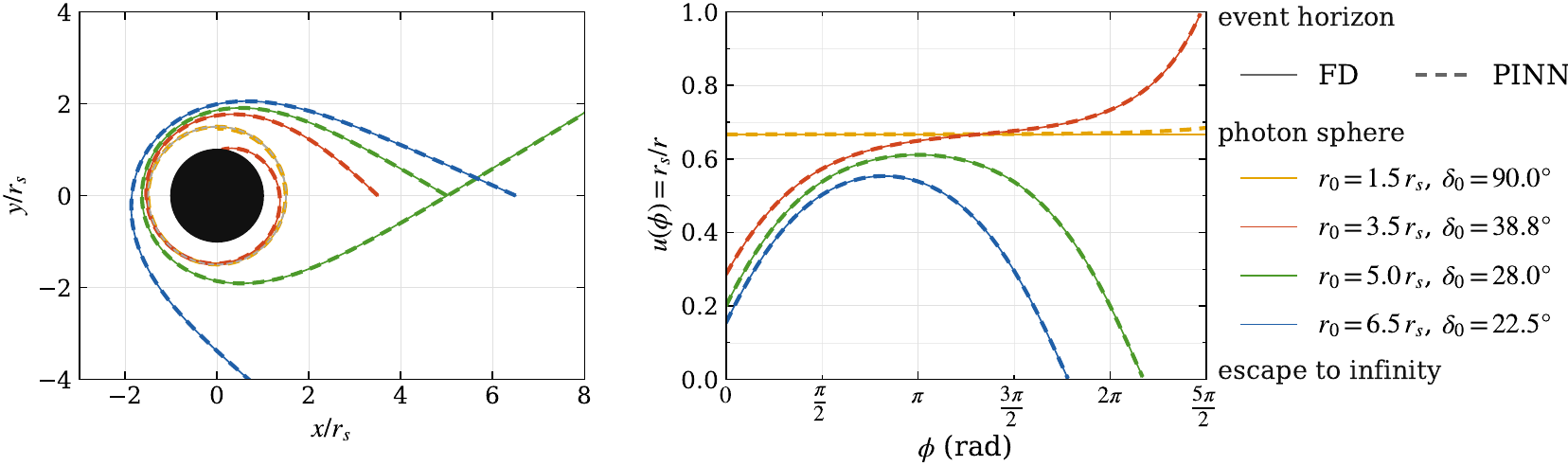}
    \caption{FD and PINN-Slicer solutions for $\Delta\phi = 0.8$, over
    $\phi \in [0, 5\pi/2]$. The photon-sphere orbit clearly diverges
    from the FD reference before $\phi = 5\pi/2$.}
    \label{fig:fdpinn080}
\end{figure}

\begin{figure}[h]
    \centering
    \includegraphics[width=\linewidth]{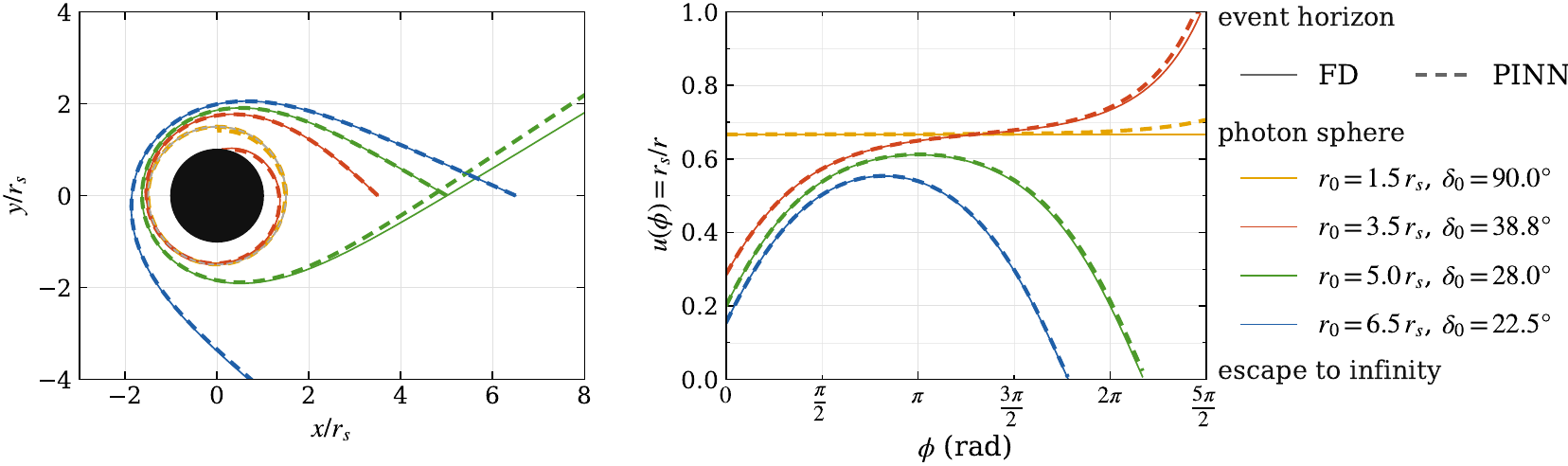}
    \caption{FD and PINN-Slicer solutions for $\Delta\phi = 1.6$. All
    orbits are visibly shifted relative to the FD reference.}
    \label{fig:fdpinn160}
\end{figure}

Figures~\ref{fig:fdpinn005}--\ref{fig:fdpinn160} show the same
comparison as Figure~\ref{fig:fdpinn-slicer010} but for PINNs trained and deployed using other $\phi$-slices. As $\Delta\phi$ increases, the recursively deployed
PINN progressively diverges from the FD reference.

\vspace{4ex}
\vspace{4ex}
\vspace{4ex}
\vspace{4ex}

\section{PINN-Slicer vs FD over a grid of initial conditions}
\label{app:rmse-vs-ic}

\begin{figure}[h]
    \centering
    \includegraphics[width=\linewidth]{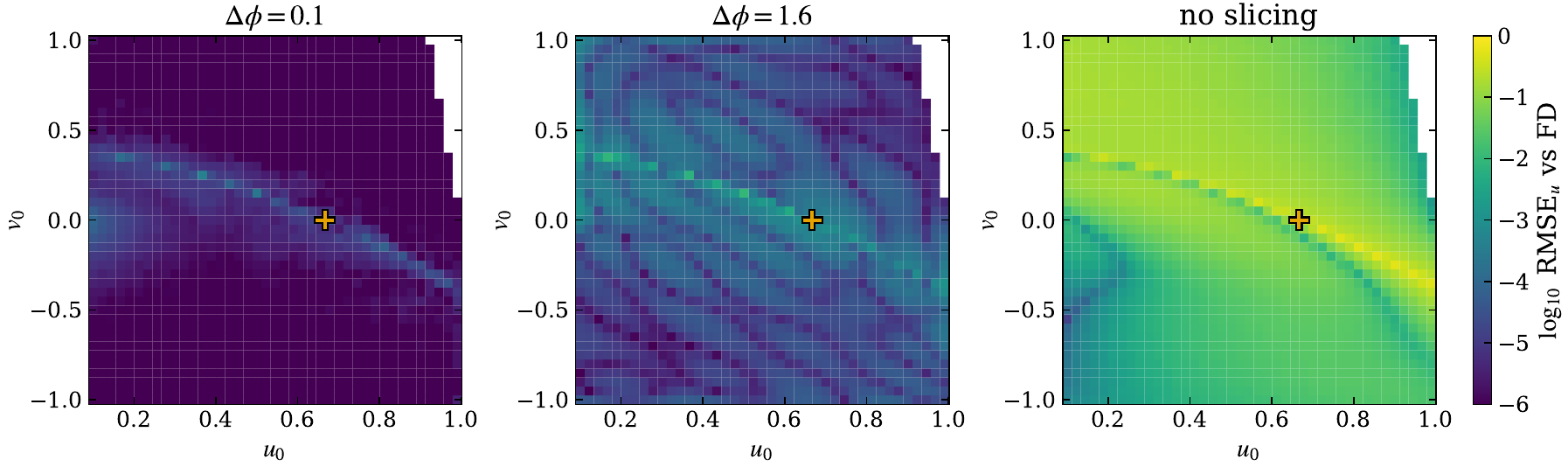}
    \caption{RMSE$_u$ between FD and the PINN trained with $\Delta\phi = 0.1$, $\Delta\phi = 1.6$ and without slicing ($\Delta\phi = 5\pi/2$), over the $41 \times 41$ grid of initial conditions. The `+' marks the photon-sphere orbit, $(u_0, v_0) = (2/3, 0)$. The white cells (upper right) correspond to photons starting close to the event horizon and moving inward: they are absorbed within a small angle $\phi$, leaving too few points to compare with the FD solution.}
    \label{fig:rmse-vs-ic}
\end{figure}

We tested PINN-Slicer over a $41 \times 41$ grid of initial conditions $(u_0, v_0)$ covering the training region. For each slice $\Delta\phi$ (0.05, 0.1, 0.2, 0.4, 0.8, 1.6), the model trained with that slice was deployed recursively and compared to the FD solution, using the RMSE of $u$ over the $\phi$ points where both solutions exist (Figure~\ref{fig:rmse-vs-ic}). The PINN without slicing is shown as well.

\begin{table}[h]
    \centering
    \caption{Median and maximum RMSE$_u$, and percentage of orbits with RMSE$_u < 10^{-3}$, over the 1,642 comparable orbits of the grid, for each $\Delta\phi$ and without slicing.}
    \label{tab:rmse-vs-ic}
    \begin{tabular}{lccc}
    \toprule
    $\Delta\phi$ & median RMSE$_u$ & max RMSE$_u$ & RMSE$_u < 10^{-3}$ (\%) \\
    \midrule
    0.05 & $6.98 \times 10^{-7}$ & $5.75 \times 10^{-4}$ & 100.0 \\
    0.1 & $4.73 \times 10^{-7}$ & $3.98 \times 10^{-4}$ & 100.0 \\
    0.2 & $6.15 \times 10^{-7}$ & $2.78 \times 10^{-4}$ & 100.0 \\
    0.4 & $8.54 \times 10^{-7}$ & $6.27 \times 10^{-4}$ & 100.0 \\
    0.8 & $1.84 \times 10^{-6}$ & $1.77 \times 10^{-3}$ & 99.9 \\
    1.6 & $3.36 \times 10^{-5}$ & $9.57 \times 10^{-3}$ & 98.8 \\
    no slicing & $4.36 \times 10^{-2}$ & $5.36 \times 10^{-1}$ & 3.3 \\
        \bottomrule
    \end{tabular}
\end{table}

The bright arc showcasing large errors corresponds to the set of initial conditions of near-critical orbits, where a slight change in initial conditions changes the outcome (absorption or escape to infinity).

For $\Delta\phi \le 0.4$, 100\% of the 1,642 comparable orbits have RMSE$_u < 10^{-3}$, with a median below $10^{-6}$ (Table~\ref{tab:rmse-vs-ic}). Without slicing, only 3.3\% of the orbits reach this accuracy.

\end{document}